\documentclass[letterpaper, 10 pt, conference]{ieeeconf}  

\IEEEoverridecommandlockouts                              

\usepackage{graphicx} 

\usepackage{amsmath}
\usepackage{amsfonts}
\usepackage{amssymb}
\usepackage{algorithm}
\usepackage[noend]{algpseudocode}
\usepackage{multirow}
\usepackage{subfiles}
\usepackage{xspace}
\usepackage{svg}

\title{\LARGE \bf
\ipimLong
}
\author{Jingyang You$^{1}$, Hanna Kurniawati$^{1}$ and Lashika Medagoda$^{2}$
\thanks{ \scriptsize $^{1}$ School of Computing, The Australian National University, Canberra ACT 2600, Australia {\tt \{jingyang.you ; hanna.kurniawati\}@anu.edu.au}}%
\thanks{ \scriptsize $^{2}$ Abyss Solutions, 11-21 Underwood Rd,
         Homebush NSW 2140, Australia {\tt lashika@abysssolutions.com.au}}%
}

\newcommand{\sref}{Section~\ref}

\newcommand{\aref}[1]{Algorithm~\ref{#1}}

\newcommand{\fref}[1]{Fig.~\ref{#1}}

\newcommand{\cpts}{$\mathcal{C}(\mathbf{q})$\xspace}
\newcommand{\ipim}{\textbf{IPIM}\xspace}
\newcommand{\tp}{\textbf{TP}\xspace}
\newcommand{\tpipim}{\tp-\ipim}
\newcommand{\ipimLong}{Inspection Planning Primitives with Implicit Models\xspace}
\newcommand{\iris}{\textbf{IRIS}\xspace}
\newcommand{\irism}{\textbf{IRIS-M}\xspace}

\begin{document}

\maketitle
\thispagestyle{empty}
\pagestyle{empty}

\begin{abstract}
The aging and increasing complexity of infrastructures make efficient inspection planning more critical in ensuring safety. Thanks to sampling-based motion planning, many inspection planners are fast. However, they often require huge memory. This is particularly true when the structure under inspection is large and complex, consisting of many struts and pillars of various geometry and sizes. Such structures can be represented efficiently using implicit models, such as neural Signed Distance Functions (SDFs). However, most primitive computations used in sampling-based inspection planner have been designed to work efficiently with explicit environment models, which in turn requires the planner to use explicit environment models or performs frequent transformations between implicit and explicit environment models during planning. This paper proposes a set of primitive computations, called \ipimLong (\ipim), that enable sampling-based inspection planners to entirely use neural SDFs representation during planning. Evaluation on three scenarios, including inspection of a complex real-world structure with over 92M triangular mesh faces, indicates that even a rudimentary sampling-based planner with \ipim can generate inspection trajectories of similar quality to those generated by the state-of-the-art planner, while using up to $70\times$ less memory than the state-of-the-art inspection planner.
\end{abstract}

\section{INTRODUCTION}
\label{intro}
\begin{figure}[!t]
  \centering
  \hspace{-0.5em}
    \framebox{
    \parbox{3.2in}{
        \centering \includegraphics[scale=0.12]{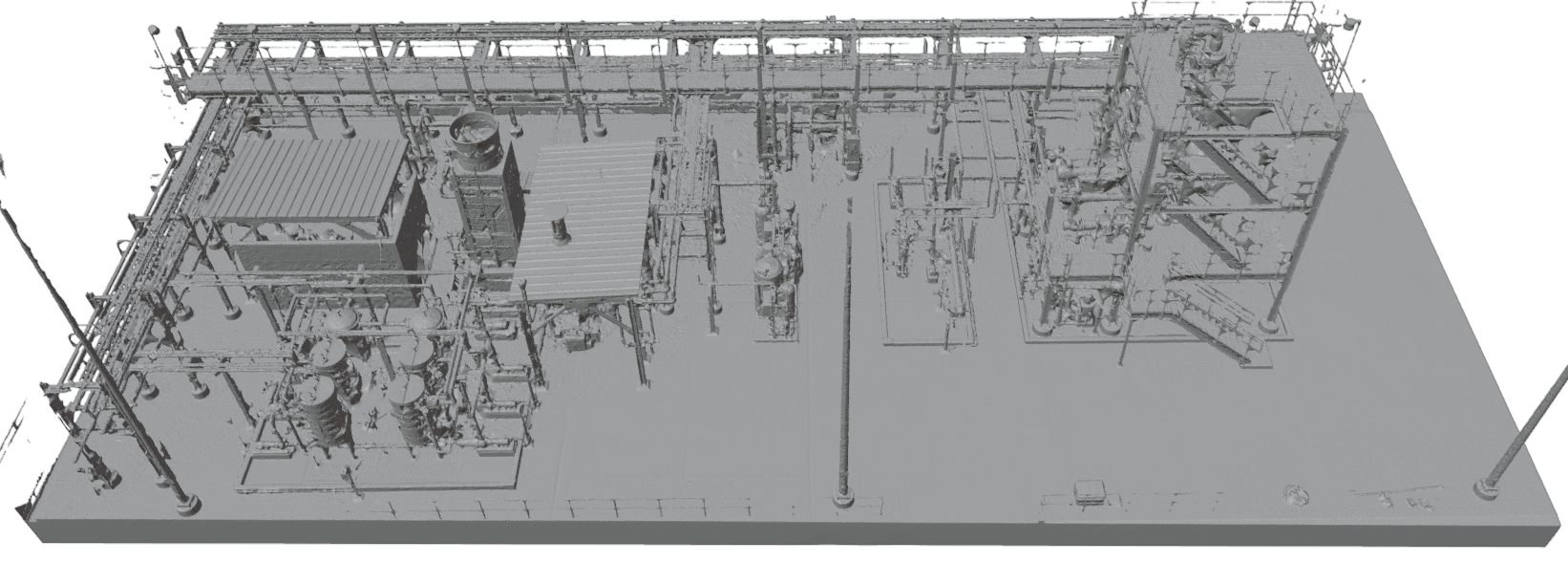}}
    }
       \caption{
       A glycol distillation plant at San Jacinto College, on which we evaluate the performance of \ipim. We name it \textbf{Plant} in \sref{exp-set}.}
  \label{fig: data}
  \vspace{-1.5em}
\end{figure}

Regular and frequent inspection of infrastructures is critical to ensure safety, while efficient inspection is important to minimize disruption.
Autonomous robots have great potential to perform such regular, frequent, and fast inspections. However, inspection planning methods that enable robots to perform autonomous inspection still face difficulties to inspect large and complex structures, such as a distillation plant illustrated in \fref{fig: data}, due to large memory requirements. This paper aims to alleviate such a difficulty. 

Inspection Planning describes the process of finding a robot's trajectory, such that upon following the trajectory, the robot perceives all points on the surfaces of the structure being inspected without any collision. Among non-myopic inspection planning methods, sampling-based inspection planning algorithms are widely used.

In sampling-based inspection planning (e.g., \cite{papadopoulos2013asymptotically, bircher2017incremental, fu2023asymptotically}), the planner samples a representative set of valid configurations, and for each of these configurations, the planner maintains information about parts of the structures perceived by the robot if it were to scan the environment from the particular configuration. This additional information about the perceived observations implies that the memory requirements of the planner increases proportionally with the number of valid sampled configurations and complexity of the environments. Moreover, in general, the number of valid samples required to generate a good inspection trajectory also increases substantially with the size and complexity of the environment being inspected. As a result, inspection planners often have prohibitively large memory requirements. Recent implicit models, such as neural Signed Distance Functions (SDFs) \cite{park2019deepsdf, wang2023co}, are known for their memory efficiency and could alleviate the mentioned memory requirement problems.

However, existing sampling-based inspection planners cannot fully benefit from such an implicit representation because existing primitives for sampling-based inspection planning were designed for explicit environment models. 
These primitives are: \textbf{(1) Collision Check} to ensure collision-free inspection trajectory, \textbf{(2) Observation Simulation} to predict observations (e.g., images, point clouds) perceived from a sampled configuration, \textbf{(3) Observation Representation} to store the simulated observation during planning and \textbf{(4) Total Coverage Check} to calculate the coverage along a potential inspection trajectory. 
Many efficient methods for each of these primitives have been proposed, but they require the environment to be represented as an explicit model.

In this paper, we propose novel primitive computations, called \ipimLong (\ipim), to enable samping-based inspection planning to leverage the more compact implicit models during the entire planning process. Specifically, 
with \ipim, the most memory-consuming components of inspection planning, i.e., explicit global environment models and local observation representations, can be 
replaced by memory-efficient neural SDFs. Although SDFs are not new, primitives that enable sampling-based inspection planners to directly use neural SDF representation 
are novel and non-trivial. 

We evaluated \ipim with a simple sampling-based inspection planner on three scenarios, including the inspection of a real-world distillation plant dataset with over 92M mesh shown in \fref{fig: data}. The results indicate that \ipim can substantially reduce the memory requirement by up to $70\times$, compared to the state-of-the-art method.

\section{Related Work\label{sec:rel}}
\subsection{Sampling-based Inspection Planning}

Inspection planning methods can be classified into myopic and non-myopic. Myopic methods generally rely on sub-modular characteristics of inspection planning problems. However, this characteristics is often false for complex cluttered environments where many areas are occluded, requiring elaborate inspection strategies that cannot be generated by myopic approaches. 


The most scalable non-myopic approach is sampling-based inspection planning. These methods compute globally feasible / optimal inspection paths via sampling. 
For instance, Random Inspection Tree Algorithm (RITA) \cite{papadopoulos2013asymptotically} is one of the first sampling-based methods that computes 
asymptotically optimal solution to inspection planning problems with non-holonomic motion constraints. 
To improve RITA's efficiency, Rapidly exploring Random Tree of Trees (RRTOT) \cite{bircher2017incremental} proposes to utilize inter-branch knowledge such that promising samples among different branches can be shared. The state-of-the-art sampling-based method today is Incremental Random Inspection-roadmap Search (IRIS) \cite{fu2023asymptotically, fu2021computationally}, which first constructs a rapidly-exploring random graph (RRG) \cite{karaman2011sampling}, then searches on the RRG for the optimal inspection path. Note, however, both RRTOT and IRIS 
require steering functions, which can be costly to compute, when the inspection is performed by non-holonomic robots.

All sampling-based inspection planners share the four primitives of inspection planning, as mentioned in \sref{intro}. Since they use explicit global environment models, they require huge memory when the structure to be inspected is large and complex, 
hindering their applicability in such inspection scenarios.

\subsection{Neural Implicit Representation}
Neural implicit representations encode 3D spatial information \cite{wang2023co} compactly in the weights of a deep neural network. \textit{Radiance Field} \cite{mildenhall2020nerf, fridovich2022plenoxels, chen2022tensorf} and \textit{Signed Distance Function} (SDF) \cite{park2019deepsdf, wang2023co, sucar2021imap} are the most widely used neural implicit representations. We focus on SDF for its high training and inference efficiency. 

SDF is a continuous function that maps any given spatial point to a 
signed distance between the point and the boundary of an object, where the sign encodes whether the point is inside (negative) or outside (positive) of the object, and an SDF value of $0$ implies the point is a surface point ---that is, the point lies on the boundary of the object \cite{park2019deepsdf}. Finding the direct $\mathbb{R}^3 \rightarrow \mathbb{R}$ SDF mapping is challenging as deep neural nets are biased towards learning low-frequency functions \cite{rahaman2019spectral}, which is not the case in real-world 3D environments where high-frequency variations in structures are common. Various methods \cite{tancik2020fourier, muller2019neural, muller2022instant} have been proposed to efficiently encode  raw coordinate inputs. These encodings enable training the SDF mapping to become nearly real-time, enabling their use in real-time robotics applications like visual SLAM \cite{wang2023co, sucar2021imap, zhu2022nice},  navigation \cite{camps2022learningdeepsdfmaps, bukhari2025differentiable} and motion planning \cite{li2024configuration}.

Despite its wide applicability in robotics, SDF is rarely used in inspection planning even though it can significantly reduce memory requirements because,  inspection planning primitives have been designed for explicit environment models. 
Removing this difficulty, we propose \ipim, a set of primitives for inspection planning with neural SDFs.

\section{Problem Definition} \label{sec:def}
We follow a typical inspection planning problem formulation, e.g., \cite{papadopoulos2013asymptotically}.  Let $C_{free}, U, \mathcal{E}$ be the collision-free part of the robot's configuration space, the control space and the explicit model of the environment, respectively. Denote the set of surface points to be inspected in $\mathcal{E}$ as $S_{\mathcal{E}}$. Our goal is to find a collision-free trajectory $\gamma^*:\left\{0, 1, \cdots, T \right\} \rightarrow C_{free}$ induced by the time-parameterized control function $\tau: \left\{0, 1, \cdots, T \right\} \rightarrow U$, such that upon following the trajectory, the robot starting from $\gamma(0)$ can perceive each point in $S$ from at least one configuration $\gamma(t), t \in \left\{0, 1, \cdots, T \right\}$, and the  workspace length of the trajectory is minimized. 
We assume the robot's geometry and kinematics are known a priori. It is equipped with visibility sensors with depth information, such as depth cameras or LiDARs. The robot performs discrete sensing, taken only at the end of each control command $\tau(t), t\in[0,T]$. 


\section{Implicit Sampling-based Inspection Planning \label{sec:met}}
\subsection{Overview}
\label{s:overview}

\begin{figure*}[t!]
    \centering
    \framebox{\includegraphics[width=0.98\textwidth]{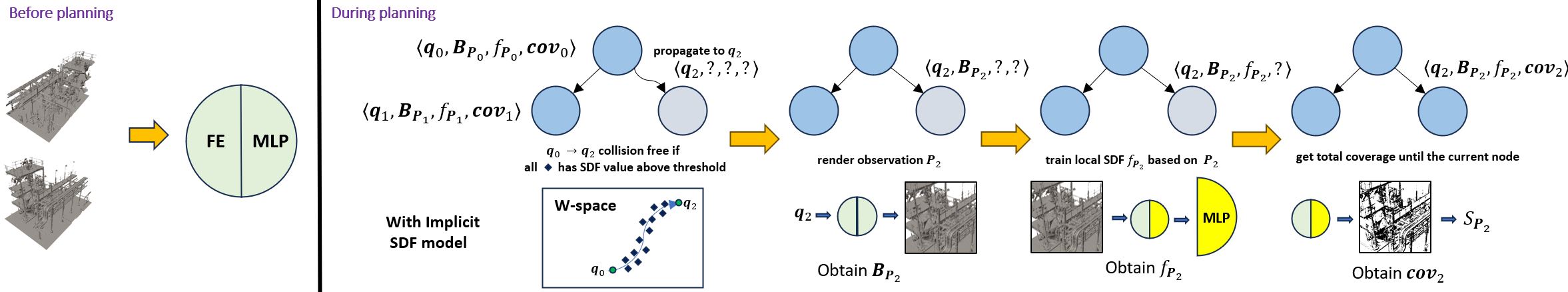}}
    \caption{Illustration of how \ipim is used with an inspection planner to reduce the memory cost. Before planning, the explicit environment $\mathcal{E}$ is converted to neural SDF $f_\mathcal{E}$ comprising a feature extractor (FE) and a Multi-Layer Perceptron (MLP). During planning, \ipim converts the four primitive computations of inspection planning with SDF to the implicit counterparts. Description about the four primitive computations is presented in \sref{intro}.}
    \label{fig: diag}
    \vspace{-1.5em}
\end{figure*}

\begin{figure}[t]
\vspace{-0.5em}
\begin{algorithm}[H]
\caption{\textbf{\tpipim} (Explicit Env. $\mathcal{E}$, Planner \tp)}\label{alg: overview}
\begin{algorithmic}[1]
\State Initialize: \tp.Tree $\mathbb{T}$
\State Initialize: Implicit SDF model $f_\mathcal{E}$ \algorithmiccomment{\sref{sdfnet}}
\While {PlanningTime=True}
    \State $\mathbf{n}_{prt}, \mathbf{n}_{chd}$, $\overline{\mathbf{n}_{prt} \mathbf{n}_{chd}}$ = \textbf{Expand($\mathbb{T}$)}
    \If {\textbf{CollisionFree}($\overline{\mathbf{n}_{prt} \mathbf{n}_{chd}}$, $f_\mathcal{E}$)} \algorithmiccomment{\sref{implicitcollision}}
        \State $\mathbf{P}$ = \textbf{SimulateObs}($\mathbf{n}_{chd}$, $f_{\mathcal{E}}$) \algorithmiccomment{\sref{implicitobs}}
        \State $\mathbf{n}_{chd}.\mathbf{B}_{\mathbf{P}} = \mathbf{P}.\mathbf{B}_{\mathbf{P}}$
        \State $f_{\mathbf{P}}$ = \textbf{ObsRepresent}($\mathbf{P}$, $f_{\mathcal{E}}$) \algorithmiccomment{\sref{localcov}}
        \State $S_{\mathbf{P}}$ = \textbf{MarchingCube($f_{\mathbf{P}})$}
        \State $\mathbf{n}_{chd}.f_{\mathbf{P}} = f_{\mathbf{P}}$
        \State $\mathbf{n}_{chd}.\textbf{cov}$ = \textbf{TotalCov}($\mathbf{n}_{chd}, S_{\mathbf{P}}$) \algorithmiccomment{\sref{globalcov}}
        \State $\mathbb{T}$.\textbf{Add}($\overline{\mathbf{n}_{prt} \mathbf{n}_{chd}}, \mathbf{n}_{chd}$)
    \EndIf
\EndWhile
\end{algorithmic}
\end{algorithm}
\vspace{-2.5em}
\end{figure}

\ipim proposes efficient primitives for inspection planning when implicit neural SDF representation is used to represent the environment and observation. Neural SDF is recognized for its rapid training and compactness, making it well-suited for planning tasks that demand memory efficiency.


With \ipim, the known explicit environment model $\mathcal{E}$ is first converted into an implicit SDF model $f_{\mathcal{E}}$ represented by a neural network (\sref{sdfnet}). Without loss of generality, we suppose \ipim is used with a sampling-based inspection planner that builds a tree $\mathbb{T} = \{ \mathbb{N}, \mathbb{E}\}$, where $\mathbb{N}$ and $\mathbb{E}$ are the set of nodes and edges in the tree. Under \ipim, each node $\mathbf{n} \in \mathbb{N}$ represents a 4-tuple $\left\langle \mathbf{q}, \mathbf{B_{P_n}},  f_\mathbf{P_n}, \textbf{cov($\mathbf{n}$)}\right\rangle$. The four elements in the tuple reflect how \ipim converts the four primitives of the planner to their implicit forms, namely:
\begin{enumerate}
     \item   \textbf{Collision Check:} Let $\mathbf{q} \in C_{free}$ be the sampled robot's configuration at node $\mathbf{n}$. Suppose the parent of $\mathbf{n}$ is $\mathbf{n}' \in \mathbb{N}$ and the configuration at $\mathbf{n}'$ is $\mathbf{q}' \in C_{free}$, then the collision check against trajectory $\overline{\mathbf{q}'\mathbf{q}}$ is performed with the implicit model $f_{\mathcal{E}}$. 
    \item  \textbf{Observation Simulation:} Let $\mathbf{P_n}$ be the raw observation perceived by the robot's configuration at node $\mathbf{n}$ (e.g., a depth image). This observation is simulated using the SDF function $f_{\mathcal{E}}$. We also derive a point cloud from $\mathbf{P_n}$. \textit{For brevity, we use the term bounding box of $\mathbf{P_n}$ to refer to the bounding box of the point cloud corresponding to the observation} $\mathbf{P_n}$. \ipim maintains the bounding box of $\mathbf{P_n}$, denoted as $\mathbf{B_{P_n}}$. \textit{In the following sections, we use $\mathbf{P}$ without the subscript $\mathbf{n}$ when referring to a general raw observation.} 
    \item \textbf{Observation Representation:} The notation $f_\mathbf{P_n}$ denotes the tiny-size multi-layer perceptron that encodes the local SDF representation of the observation $\mathbf{P_n}$. The set of local surface points, denoted as $S_\mathbf{P_n}$, can then be \textit{generated} from $\mathbf{B_{P_n}}$ and $f_\mathbf{P_n}$ with marching cube \cite{lorensen1998marching}. 
    \item \textbf{Total Coverage Check:} The notation \textbf{cov($\mathbf{n}$)} refers to the accumulated coverage in the path from the root of $\mathbb{T}$ until node $\mathbf{n}$ of $\mathbb{T}$. This coverage is calculated incrementally and implicitly as $\mathbb{T}$ being expanded. 
\end{enumerate}
The details of each of the primitives above are presented in \sref{implicitcollision}--\sref{globalcov}, respectively.

\aref{alg: overview} presents an overview of how a sampling-based inspection planner with tree representation can use \ipim and neural SDF model throughout its entire planning process. \fref{fig: diag} provides a summary of the four primitives in \ipim and how they are applied in a tree-based sampling-based inspection planner. 

\subsection{SDF Neural Network Structure \label{sdfnet}}
\ipim replaces the memory-consuming explicit environment model $\mathcal{E}$ with an implicit neural SDF, denoted as $f_{\mathcal{E}}$, that approximates the ground truth SDF. To this end, it parameterizes the neural SDF function with additional parameters that are the weights of the neural network. Specifically:
\begin{align}
    f_{\mathcal{E}}(\mathbf{x}; \boldsymbol{\theta}_1, \boldsymbol{\theta}_2) = 
    \text{MLP}\left(\text{FE}(\mathbf{x}; \boldsymbol{\theta}_1);\boldsymbol{\theta}_2 \right)
\end{align}
where $\mathbf{x} \in {\mathbb{R}}^3$ is a 3D point, and   $\boldsymbol{\theta}_1 \in \mathbb{R}^{d_1}, \boldsymbol{\theta}_2 \in \mathbb{R}^{d_2}$ are the weights of the neural network, where $d_1, d_2$ are hyperparameters determining the size of the weight.
MLP stands for \textbf{M}ulti-\textbf{L}ayer \textbf{P}erceptron. Whilst the notation $\text{FE}(\mathbf{x}; \boldsymbol{\theta}_1) = \left[ \text{HashGrid}(\mathbf{x}; \boldsymbol{\theta}_1)^T~ \text{OneBlob}(\mathbf{x})^T\right]^T$ refers to a \textbf{F}eature \textbf{E}xtractor that concatenates features obtained from \text{HashGrid} -- multiresolution hash encoding \cite{muller2022instant}, and \text{OneBlob} encoders \cite{muller2019neural}. 
The former learns spatial features from $\mathbf{x}$ and the latter describes $\mathbf{x}$ with multiple frequency bands, suitable for structures with complex geometry. 


HashGrid 
disentangles the task of mapping into two sub-tasks: \textit{(1) feature extraction} and \textit{(2) mapping from extracted features to SDF values}. Task (1) requires many parameters as it contains high-level spatial information of the environment, while task (2) usually requires only MLPs of tiny sizes. Since $\mathcal{E}$ is a priori known, we can pre-train FE ($\boldsymbol{\theta}_1$) to complete task (1). During planning, all task (1) parameters are \textit{shared} among all nodes in the tree, while the local observations in each node is fully controlled by tiny-sized parameters in downstream MLPs and parameterized by $\boldsymbol{\theta}_2$.

When training the network on the entire environment $\mathcal{E}$, we apply the mean squared loss function 
\begin{align}
    \mathcal{L}_{\mathcal{E}}(\mathbf{X}, \mathbf{Y}; \boldsymbol{\theta}_1, \boldsymbol{\theta}_2) = \frac{1}{\vert \mathbf{X} \vert} \left\Vert f_{\mathcal{E}}(\mathbf{X}; \boldsymbol{\theta}_1, \boldsymbol{\theta}_2) - \mathbf{Y}\right\Vert_2^2
\end{align}
where $\mathbf{X}, \mathbf{Y}$ are collections of 3D coordinates and corresponding SDF values obtained from the explicit representation of $\mathcal{E}$, respectively. $\vert \cdot \vert$ and $\Vert \cdot \Vert_2$ are the cardinality and the $l_2$ norm. 
We apply the truncated SDF (TSDF) \cite{park2019deepsdf} strategy, which ensures the magnitude of SDF value cannot be greater than a given threshold, to better approximate the depth values and reduce variance in points far from any surface. TSDF is applied to the ground truth SDF and to the learnt SDFs. 

\subsection{Collision Check with Implicit SDF \label{implicitcollision}}
To check whether the robot at configuration $\mathbf{q}$ is in collision or not, we first represent the robot's geometry with a sufficiently dense set of control points \cite{ramp} that encapsulates the robot, denoted by \cpts.  
For any $\mathbf{p} \in$ \cpts and threshold value $\xi > 0$,  if $f_{\mathcal{E}}(\mathbf{p}) > \xi$, we know $\mathbf{p}$ is $\xi$ units away from its closest surface. When all points in \cpts have SDF values greater than $\xi$, we can infer the robot at configuration $\mathbf{q}$ is collision free. 

\subsection{Generating Simulated Observation from SDF \label{implicitobs}}
\ipim provides a function to simulate observations from neural SDF environment model. 
Specifically, \ipim simulates observation depth image $\mathbf{P}$ from the neural SDF $f_\mathcal{E}$, instead of the explicit environment $\mathcal{E}$. To this end, \ipim marches rays emitted from a simulated depth sensor. For each ray, \ipim finds the first surface point of $\mathcal{E}$  hit by the ray, i.e., the first 
point $\mathbf{p}$ along the ray that satisfies $f_\mathcal{E}(\mathbf{p}) = 0$.
The simulated depth provided by the ray is then the distance the ray travels to reach $\mathbf{p}$.



\subsection{Representing Simulated Observation \label{localcov}}

Instead of storing the simulated observation explicitly, 
\ipim represents them as local SDF, denoted as  $f_{\mathbf{P}}$ and encoded as 
a tiny-sized local MLP that reuses the feature extractor FE of $f_{\mathcal{E}}$.

A key functionality of the simulated observation is to estimate the coverage of different paths. Suppose $\mathbf{P_n}$ is the depth image perceived from a robot's configuration at node $\mathbf{n}$ of the plannng tree $\mathbb{T}$. 
We define coverage of $\mathbf{n}$ as $\#\text{Covered($\mathbf{P_n}$)} / |S_{\mathcal{E}}|$, where $\#$Covered($\mathbf{P_n}$) and $|S_{\mathcal{E}}|$ are the number of surface points of $\mathcal{E}$ visible in $\mathbf{P_n}$ and the total surface points, respectively. The set of surface points of $\mathcal{E}$ visible in $\mathbf{P_n}$ can be obtained by performing marching cube \cite{lorensen1998marching} on the SDF model $f_{\mathbf{P_n}}$. Modern libraries like \textit{Pytorch3D} \cite{ravi2020pytorch3d} support GPU-accelerated marching cube. These libraries enable marching cube operation to run fast, especially when $\mathcal{E}$ has a high resolution. Moreover, by using marching cube, the explicit $\mathbf{P}$ or equivalently the corresponding point cloud, can be discarded, and only the weights of $f_{\mathbf{P_n}}$, which is tiny, need to be maintained, thereby reducing memory usage. Since coverage is computed for each node of $\mathbb{T}$, the total memory \ipim saves is substantial. 

Now the question is how to efficiently learn an accurate $f_{\mathbf{P}}$ from $\mathbf{P}$, so that the implicit $f_{\mathbf{P}}$ can be used faithfully just like the explicit $\mathbf{P}$. 
Such an accurate $f_{\mathbf{P}}$ also minimizes labeling unvisited areas as visited or vice versa, helping to compute an accurate estimate of paths' coverage.  
Training $f_{\mathbf{P}}$ from scratch generally requires more memory or time than storing $\mathbf{P}$ explicitly. However, since  $\mathcal{E}$ is static and known a priori, any $\mathbf{P}$ must correspond to a certain part of $\mathcal{E}$. 
Since FE of $f_\mathcal{E}$ contains the spatial information of the entire $\mathcal{E}$, it also contains the area corresponding to any $\mathbf{P}$. 
Therefore, FE of $f_\mathcal{E}$ can be reused when encoding $\mathbf{P}$. The local SDF of $\mathbf{P}$ is hence $f_{\mathbf{P}} = \text{MLP}\left(\text{FE}(\mathbf{x}; \boldsymbol{\theta}_1);\boldsymbol{\theta}_{\mathbf{P}} \right)$, where the FE of $f_\mathcal{E}$ is reused, and the parameters of $\boldsymbol{\theta}_{\mathbf{P}}$ are re-trained and maintained. 

Before defining the loss function used to train $f_{\mathbf{P}}$, let us first define our accuracy goals:  
(1) surface points visible from $\mathbf{P}$ should have SDF values close to $0$, and (2) invisible (occluded / uncovered) surface points should have large TSDF magnitudes so that their corresponding vertices disappear in the mesh obtained from $f_{\mathbf{P}}$. The first objective can be solved by minimizing $\mathcal{L}_{\mathcal{E}}$, while (2) cannot, as the shared FE has dominantly more parameters than that of MLP. If only $\mathcal{L}_{\mathcal{E}}$ is minimized, FE \textit{predicts} accurate SDF values on areas not visible in $\mathbf{P}$, which is undesirable. Instead, we want the SDF of areas not visible in local observation $\mathbf{P}$ to be large in magnitude, so that the marching cube \textit{does not} generate surface points in areas not visible in $\mathbf{P}$. 

Let us define a ray as $\mathbf{r} = \mathbf{o} + t \cdot \mathbf{d}$, where $\mathbf{o, d}$ are the ray origin and direction, respectively, and $t$ is the distance traveled along $\mathbf{d}$. From the depth information in $\mathbf{P}$, we know when the ray hits a surface, say at $t = t_0$, all subsequent points on the ray, i.e., at $t=t_0 + \delta$ where $\delta > 0$, are occluded.
For learning the neural SDF, the occluded points have target TSDF values $\min \left\{tr, \delta\right\}$ where $tr > 0$ is the already set truncation value for TSDF. These occluded points will also be fed into the network to weaken the dominance of FE, letting MLP predict truncation values for occlusions. The loss function for learning local SDF is then:
\begin{multline}
    \mathcal{L}_{local} (\mathbf{X}_{vis}, \mathbf{X}_{occ}, \mathbf{Y}_{vis}, \mathbf{Y}_{occ}; \boldsymbol{\theta}_1, \boldsymbol{\theta}_\mathbf{P}) \\
    = \lambda_{vis} \cdot \mathcal{L}_{\mathcal{E}}(\mathbf{X}_{vis}, \mathbf{Y}_{vis}; \boldsymbol{\theta}_1, \boldsymbol{\theta}_\mathbf{P}) \\
    + \lambda_{occ} \cdot \frac{1}{\vert \mathbf{X}_{occ} \vert} \left\Vert f_\mathcal{E}(\mathbf{X}_{occ} ; \boldsymbol{\theta}_1, \boldsymbol{\theta}_\mathbf{P}) - \mathbf{Y}_{occ} \right\Vert_2^2
\end{multline}
where $\lambda$s are hyperparameters, $vis,~occ$ stand for visible and occluded. Note the weight of FE, $\boldsymbol{\theta}_1$, is frozen during learning MLP weights $\boldsymbol{\theta}_\mathbf{P}$. The pseudocode for learning $f_\mathbf{P}$ is shown in \aref{alg:local_learner}. This strategy enables inspection planners to be much more efficient in representing locally observed environments.

\begin{figure}[t]
    \vspace{-0.5em}
    \begin{algorithm}[H]
    \caption{\textbf{ObsRepresent} (Observation $\mathbf{P}$, TSDF $f_{\mathcal{E}}$)}\label{alg:local_learner}
    \begin{algorithmic}[1]
    \State Initialize: New multi-layer perceptron MLP
    \State Initialize: TSDF model $f_{\mathbf{P}} = \text{MLP} \circ  f_{\mathcal{E}}.\text{FE}  $
    \State Initialize: Freeze $\mathbf{\boldsymbol \theta}_1$, weight of $f_{\mathcal{E}}.\text{FE}$
    \State Initialize: Reset $\mathbf{\boldsymbol \theta}_\mathbf{P}$, weight of $\text{MLP}$
    \State Initialize: Ray origin, directions $\mathbf{O}$, $\mathbf{D}$ from $\mathbf{P}$
    \State Initialize: Target depth $\mathbf{T}$  from $\mathbf{P}$, truncation $tr=f_{\mathcal{E}}.tr$
    \State Sample $\mathbf{T}_{vis}$ in $[\mathbf{T} - tr, \mathbf{T}]$ 
    \State Sample $\mathbf{T}_{occ}$ in $[\mathbf{T}, \mathbf{P}.\text{MaxDepth}]$ 
    \State $\left\{\mathbf{X}_{vis}, \mathbf{X}_{occ}\right\} = \mathbf{O} + \left\{\mathbf{T}_{vis}, \mathbf{T}_{occ}\right\} \odot \mathbf{D}$
    \State $\mathbf{Y}_{vis}= \mathbf{T} - \mathbf{T}_{vis}$ ; \; $\mathbf{Y}_{occ} = \min \left\{\mathbf{T} - \mathbf{T}_{occ}, -tr \right \} $
    \While{Iter $<$ MaxIter}
    \State $\boldsymbol{\theta}_\mathbf{P} \leftarrow$ $\boldsymbol{\theta}_\mathbf{P} - \alpha \nabla_{\boldsymbol{\theta}_\mathbf{P}} \mathcal{L}_{local}$
    \EndWhile
    \State \textbf{Return} $f_{\mathbf{P}}$
    \end{algorithmic}
    \end{algorithm}
    \vspace{-2.5em}
\end{figure}

\begin{figure}
\vspace{-0.5em}
\begin{algorithm}[H]
\caption{\textbf{TotalCov} (Node $N$, Surface points $S_{\mathbf{P}_{N}}$)}\label{alg:total_cov}
\begin{algorithmic}[1]
\State Initialize: Predecessor Nodes ${1, 2, \cdots, N}$
\State Initialize: Bounding boxes $\{\mathbf{B}_{\mathbf{P}_i}\}_{i=1}^{N}$, Tolerance $\epsilon > 0$
\For{i \textbf{in} $\{1, 2, \cdots, N - 1\}$}
    \If{$\mathbf{B}_{\mathbf{P}_i} \cap \mathbf{B}_{\mathbf{P}_N} \ne \varnothing$}
    \For{$\mathbf{p} \in S_{\mathbf{P}_N}$}
        \If{$\mathbf{p} \in \mathbf{B}_{\mathbf{P}_i}~\textbf{and}~|f_{\mathbf{P}_j}(\mathbf{p})| < \epsilon$  }
            \State Remove $\mathbf{p}$ from $S_{\mathbf{P}_N}$
        \EndIf
    \EndFor
    \EndIf
\EndFor
\State \textbf{Return} $|S_{\mathbf{P}_N}| / |S_\mathcal{E}|$ + ($N-1$).\textbf{cov}
\end{algorithmic}
\end{algorithm}
\vspace{-3em}
\end{figure}

\subsection{Total Coverage Checking \label{globalcov}}
Each path from the root to a leaf node in the planning tree $\mathbb{T}$ forms an inspection trajectory. The total coverage along trajectories can be used as termination criterion and planning heuristic. However, unlike explicit representations where the total coverage can be easily retrieved from set unions, 
trivially concatenating SDFs in nodes along trajectories do not result in another SDF that describes the total coverage.

To perform \textbf{Total Coverage Check} along a trajectory,  \ipim requires the set of local surface points of the nodes in the trajectory.  
The local surface point $S_{\mathbf{P_n}}$ of node $\mathbf{n}$ can be obtained by performing marching cube on the learnt local SDF $f_{\mathbf{P_n}}$, assuming the bounding box of the observation $\mathbf{B}_{\mathbf{P_n}}$ is defined. 
To ensure $S_{\mathbf{P_n}}$ is a subset of global surface points, so that no extra surface points are generated, the bounding box $\mathbf{B}_{\mathbf{P_n}}$ must be snapped to the grid used by the marching cube on $f_\mathcal{E}$. In practice, one only needs to snap the two extreme points of $\mathbf{B}_{\mathbf{P_n}}$ to the grid.
Note that using the above method, the set of local surface points of a node is only needed the first time coverage is computed for the particular node, and can be discarded immediately after, to keep the memory consumption low.

Suppose a trajectory has $N - 1$ nodes with local SDFs $\left\{f_{\mathbf{P}_i}\right\}_{i=1}^{N-1}$ and total coverage $C$. When a new node $N$ is added to the trajectory, we first identify the number $K$ of newly visited surface points at the new node $N$ ---that is, surface points that do not intersect with observations in the predecessor nodes $\{1, 2, \cdots, N-1 \}$. Then, the total coverage for the new trajectory, consisting of $\{\text{node-$1$}, \cdots, \text{node-$(N-1)$}, \text{node-$N$}\}$, can be calculated incrementally as $C + K / |S_{\mathcal{E}}|$. 

To compute the above coverage, we need an efficient method to check if a surface point in the newly added node, $\mathbf{p} \in S_{\mathbf{P}_N}$, intersects with an observation in the predecessor nodes $\left\{\mathbf{P}_i\right\}_{i=1}^{N-1}$.  
To this end, first, $\mathbf{B}_{\mathbf{P}_N}$ is checked against $\left\{\mathbf{B}_{\mathbf{P}_i}\right\}_{i=1}^{N-1}$. 
If $\mathbf{B}_{\mathbf{P}_N}$ and $\mathbf{B}_{\mathbf{P}_i}$ do not intersect, then $S_{\mathbf{P}_N} \cap S_{\mathbf{P}_i} = \varnothing$ and hence $\mathbf{p} \notin S_{\mathbf{P}_i}$.
If the bounding box $\mathbf{B}_{\mathbf{P}_N}$ and $\mathbf{B}_{\mathbf{P}_i}$ intersects, we need to check if $\mathbf{p} \in S_{\mathbf{P}_i}$. This check can be done efficiently by first checking if $\mathbf{p}$ is in $\mathbf{B}_{\mathbf{P}_i}$. 
If it is, then $\mathbf{p} \in \mathbf{S}_{\mathbf{P}_i}$ if $|f_{\mathbf{P}_i}(\mathbf{p})| < \epsilon$ where $\epsilon > 0$ is a threshold to check if the SDF is close to 0. 
The number $K$ of newly visited points  is then the cardinality of the set $\{ \mathbf{p} \in S_{\mathbf{P}_N} | ~\forall i \in [1, N - 1], \mathbf{p} \notin  \mathbf{S}_{\mathbf{P}_i} \}$,
and the total coverage for the trajectory consisting of $\{\text{node-$1$}, \cdots, \text{node-$(N-1)$}, \text{node-$N$}\}$ is $C + K / |S_{\mathcal{E}}|$. \aref{alg:total_cov} describes the method to fuse local SDFs to calculate the total coverage. 

By converting all primitives of inspection planning to work directly with implicit environment models, \ipim enables the development of novel memory-efficient inspection planning.

\section{Experiments \label{sec:exp}}
\subsection{Experiment Settings}
\label{exp-set}

\begin{figure}[t]
  \centering
  {
      \begin{minipage}{0.48\textwidth}
      \hspace{-0.5em}
        \centering
        \framebox{\parbox{3.28in}{\centering \includegraphics[scale=0.15]{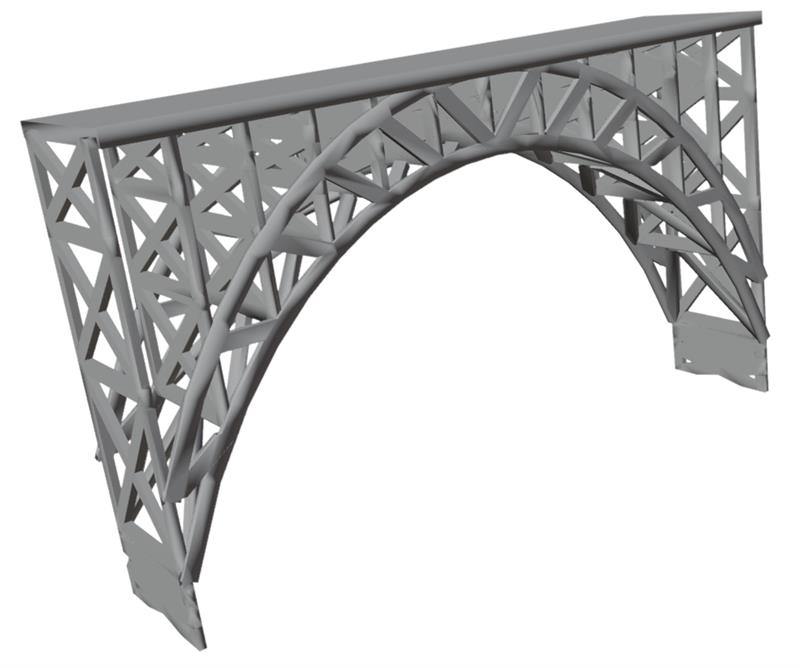} \hspace{1.5em}\includegraphics[scale=0.09]{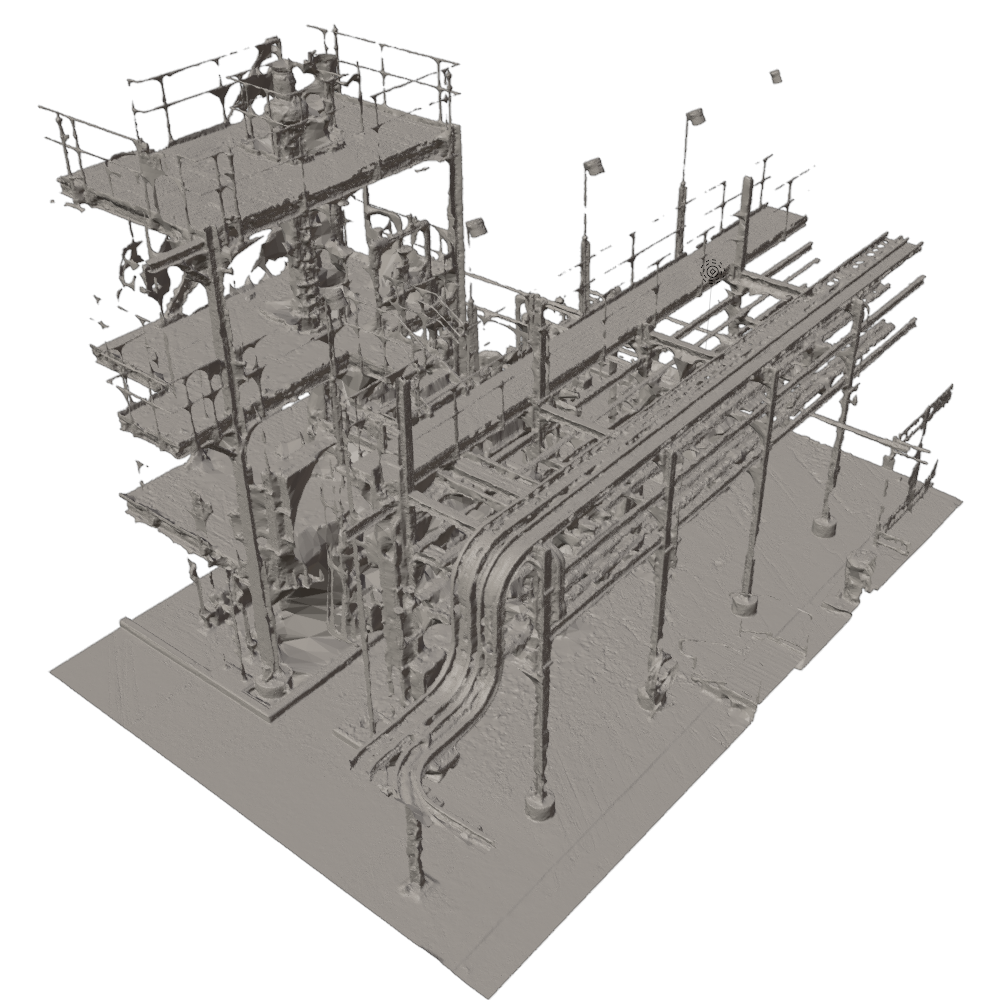}}}
      \end{minipage}
       \caption{
           From left to right: \textbf{Bridge} taken from \cite{fu2023asymptotically}, and \textbf{Plant-s} taken as a subset of \textbf{Plant} shown in \fref{fig: data}. \label{fig: bridge_abyss}}
   }
  \vspace{-1.5em}
  
\end{figure}

We evaluate our proposed method on three scenarios with screenshots of the mesh shown in \fref{fig: data} and \fref{fig: bridge_abyss}:
\begin{enumerate}
    \item \textbf{Bridge} has 7.4K mesh vertices and 9.4K mesh faces, taken from \cite{fu2023asymptotically, fu2021computationally}. We have reduced the scale of this environment by a factor of ten, so it has size 6m $\times$ 2m $\times$ 4m.
    \item \textbf{Plant} refers to a glycol distillation plant at the San Jacinto College. The mesh model of the plant is obtained by processing raw scans collected from a Leica RTC360 Laser Scanner and a BLK ARC Lidar scanner. This scenario consists of 52.5M vertices and 92M faces, and is sized 44m $\times$ 19m $\times$ 15m.
    \item \textbf{Plant-s} is a subset of \textbf{Plant} with size 15m $\times$ 10m $\times$ 10m, consisting of 
    10.4M vertices and 20.9M faces. 
\end{enumerate}

The robot used for inspection is a drone with 10cm side length. Its C-space has 5 dimensions, consisting of the position of the drone's center of mass, yaw, and pitch angle. The drone is equipped with a visibility sensor with limited range (1m for the \textbf{Bridge} scenario and 3m for the other scenarios) and $90^\circ$ horizontal and vertical field-of-view. 

\ipim works as a framework to be combined with an inspection planner. To demonstrate its performance, we propose a simple tree-based planner \tp to be used with \ipim. \tp itself is implemented on the CPU, while \ipim is accelerated by the GPU. The pseudocode of \tp with \ipim is shown in \aref{alg: overview} with modification in the tree expansion, and additional prunings. When expanding the planning tree, inspired by \cite{rita_constraint}, we bias \tp towards nodes with higher coverage, so that  $Pr(\textbf{n} ~\text{being sampled}) \propto 1/N_d(\textbf{n}) + \alpha \textbf{n}.\textbf{cov}$ with constant $\alpha > 0$, $N_d(\textbf{n})$ being the number of nodes whose configuration to \textbf{n} is less than $d$ units. The expansion is performed single-threaded, such that only one new node is expanded every time. \tp also prunes the tree every certain iterations (2000 in all our experiments). The pruning maintains only the branch with the highest coverage. To emphasise, the main reason of performing pruning is for \tp to achieve a higher coverage. Without pruning, \tp can be viewed as a RITA \cite{papadopoulos2013asymptotically} inspection planner that fails to achieve a decent coverage in large complex scenarios, as suggested by \cite{rita_constraint} and tested by the authors. Note that \tp is a simple example planner used to test the performance of \ipim and the pruning step will certainly not harm the per-node memory reduction of \ipim as \tp and \ipim are completely separate and independent.


We compare the performance of four methods: \textbf{\tp with \ipim (\tpipim)}, \textbf{\tp without \ipim (\tp)}, \textbf{\iris  \cite{fu2023asymptotically}} --- a state-of-the-art sampling-based inspection planner, and \textbf{\irism}. \irism is our modification \iris where the visibility set computation is parallelized by \textit{Open3D} \cite{Zhou2018} with 12 CPU threads. Without this modification, \iris failed in \textbf{Plant-s} and \textbf{Plant}. We use the same library and the same number of CPU threads for \tp. Note that although \tp and \tpipim are tree-based while \iris and \irism are graph-based, they are still comparable as they all share the same four primitive computations of inspection planning.



\begin{table*}[t]

\def\arraystretch{1.1}
\caption{Coverage, Cost, and Memory w.r.t. Environments and Algorithms}
\label{tbl: all}
\addtolength{\tabcolsep}{-3.2pt}
\begin{center}
\vspace{-1.3em}
\begin{tabular}{|c||c|c|c|c||c|c|c|c||c|c|c|c||}
\hline
\textbf{Environment} & \multicolumn{4}{c||}{\textbf{Bridge (30 minutes)} }  & \multicolumn{4}{c||}{\textbf{Plant-s (2 hours)} }  & \multicolumn{4}{c||}{\textbf{Plant (2 hours)} }  \\
\hline 
\textbf{Planner} & \iris & \irism & \tp &  \tpipim & \iris & \irism & \tp &  \tpipim & \iris & \irism & \tp &  \tpipim\\
\hline
\textbf{Mem Base (MB)} & $0.8 \pm 0.0$ & $0.8 \pm 0.0$ & \textbf{0.5 $\pm$ 0.0} &  $3.6 \pm 0.0$ & N.A. & $795 \pm 10$& $930 \pm 17$ & \textbf{32 $\pm$ 0} & N.A. & $2178 \pm 15$&  $2179 \pm 26$ & \textbf{31 $\pm$ 0}\\
\hline
\textbf{Mem Total Limit} & \multicolumn{12}{c||}{Each of \iris, \irism, and \tp: \textbf{96 GB}; \tpipim: \textbf{\textbf{1}} \textbf{GB}}  \\
\hline
\textbf{\# Out of Mem} & 0/10 & 0/10 & 0/10 & 0/10 & N.A. & 6/10 & \textbf{0/10} & \textbf{0/10} & N.A. & 10/10 & \textbf{0/10} & \textbf{0/10}   \\
\hline
\textbf{Coverage ($\%$)} & $55 \pm 1$ & $55 \pm 1$ & \textbf{88 $\pm$ 0} & $54 \pm 3$& N.A. & $56 \pm 3$& \textbf{84 $\pm$ 1} & $54 \pm 3$& N.A. & $21 \pm 2$& $19 \pm 2$& \textbf{24 $\pm$ 2}\\
\hline
\textbf{Cost} & \textbf{348 $\pm$ 6} & \textbf{348 $\pm$ 6} & $1965 \pm 65$& $629 \pm 29$& N.A. & \textbf{159 $\pm$ 2} & $361 \pm 17$ & $221 \pm 13$& N.A. & $286 \pm 14$& \textbf{188 $\pm$ 13}& $328 \pm 20$\\
\hline
\end{tabular}
\end{center}
\vspace{-2.5em}

\end{table*}

\tpipim requires an implicit model of the environment, $f_{\mathcal{E}}$. To train $f_{\mathcal{E}}$, we uniformly sample points in both near-surface and far-surface areas. Afterwards, TSDF values of those points can be evaluated from the explicit environment $\mathcal{E}$. The loss function used to encode $\mathcal{E}$, $\mathcal{L}_{\mathcal{E}}$, can then be optimized and $f_{\mathcal{E}}$ is obtained. \iris and \irism require manually setting the number of nodes in their RRGs where they search for the optimal inspection path. To set this parameter, we performed preliminary runs to find a suitable parameter for each scenario, which turned out to be 1000 nodes. To set other hyperparameters in \iris and \irism, we follow \cite{fu2023asymptotically, fu2021computationally}.

We use 2.8GHz CPUs and a single RTX3090 GPU for all experiments. \tpipim and \tp are implemented using Python for easier deep learning implementation using PyTorch \cite{pytorch}. We use Pytorch3D \cite{ravi2020pytorch3d} to accelerate the marching cube algorithm of \tpipim with GPU. 
 \iris and \irism are implemented in C++, following the official implementation. 
\iris uses spherical sectors as their visibility sets. We change them to pyramids to align with the common assumption, so that all four methods to compare share the same geometry of visibility set. The code of both \iris and \irism are taken and modified from the official implementation of \cite{fu2023asymptotically, fu2021computationally}. C++ is more memory efficient than Python, giving \iris and \irism the upper hand. 

Details related to training the environment model $f_{\mathcal{E}}$ and local SDFs $f_\mathbf{P}$ are as follows. $f_{\mathcal{E}}$ takes $\sim5$ minutes to train with a total of 426K parameters for \textbf{Bridge} and 6.9M parameters for \textbf{Plant-s} and \textbf{Plant}. During planning, $f_\mathbf{P}$ are 3-layer MLPs with 505 parameters for \textbf{Bridge} and 761 parameters for \textbf{Plant-s} and \textbf{Plant}. Each $f_\mathbf{P}$ is trained for 20 iterations in \textbf{Bridge} and 100 iterations in \textbf{Plant-s} and \textbf{Plant}, with learning rate 3e-3, using AdamW \cite{loshchilov2019decoupledweightdecayregularization} optimiser. 
\tpipim plans approximately 8-9 nodes per second in the most complex \textbf{Plant} scenario, due to the small size of $f_\mathbf{P}$.
\begin{figure}[t!]
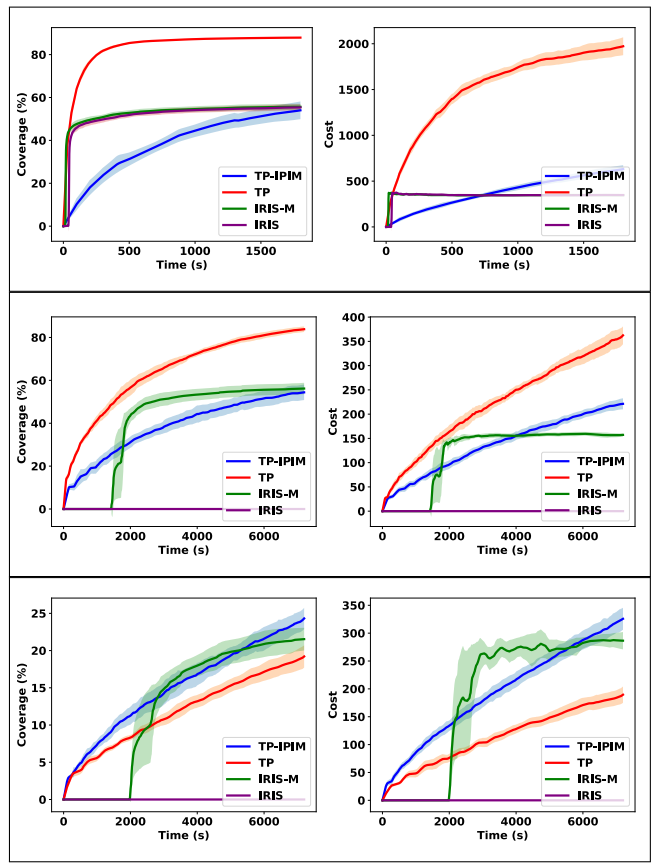

  \centering
  \vspace{0.7em}
  \framebox[0.48\textwidth]{

     \parbox{3.4in}{\includesvg[width=0.48\textwidth]{imgs/bridge.svg}} 
     }
    \framebox[0.48\textwidth]{

     \parbox{3.4in}{\includesvg[width=0.48\textwidth]{imgs/abyss.svg}} 
     }
        \framebox[0.48\textwidth]{

     \parbox{3.4in}{\includesvg[width=0.48\textwidth]{imgs/full.svg}}  
     }
  \caption{From left to right: planning time v.s. inspection coverage, and planning time v.s. cost of the inspection path. From up to bottom: scenario \textbf{Bridge}, scenario \textbf{Plant-s} and scenario \textbf{Plant}.}
  \label{fig: cov}
  \vspace{-2em}
\end{figure} 

\subsection{Performance Comparison}
In this section, we compare the performance among \tpipim, \tp, \iris and \irism, in terms of coverage, cost, and baseline memory and total memory consumption. Coverage is as defined in {\sref{localcov}}.
The cost is defined as the length of the inspection trajectory, and the baseline memory is counted as the size of the explicit (\iris, \irism, \tp) / implicit (\tpipim) global environment model, plus the size of all local observation representations in planning nodes. As discussed in \sref{intro}, these two components are necessary to most of the sampling-based inspection planning algorithms, and contribute most to the total memory consumption. We emphasize that even if we used GPU in \tpipim, the storage of planning nodes (e.g., MLP weights $f_{\mathbf{P}}$) is moved to CPU immediately after the GPU calculation is complete.

All planners are run for 10 times in three scenarios. We set the planning time limit (which doesn't include the time of obtaining $f_{\mathcal{E}}$ and I/O) to be 30 minutes, 2 hours and 2 hours for \textbf{Bridge}, \textbf{Plant-s} and \textbf{Plant}, respectively.  We set the total memory limit to be 96 GB for \iris, \irism, \tp and only 1 GB for \tpipim, and record the number of out-of-memory runs. Detailed results are shown in Table. \ref{tbl: all} and Figure. \ref{fig: cov}.

  \textbf{1) Comparison between \tp and \tpipim:} In \textbf{Bridge} and \textbf{Plant-s} scenarios, \tp plans faster than \tpipim, as \tpipim trains local MLP online to represent the local observation. 
  Though in \textbf{Plant}, \tpipim is faster than \tp. With a complex environment, the explicit representation of local observations is slow even with parallelized computations using modern libraries. 
  In \textbf{Plant}, which is the largest scenario, with \ipim, the same planner can be sped-up with substantially improved memory efficiency.

\begin{figure}[t!]
  \centering
  \framebox[0.48\textwidth]{

     \parbox{5in}{\centering \includegraphics[width=0.23\textwidth]{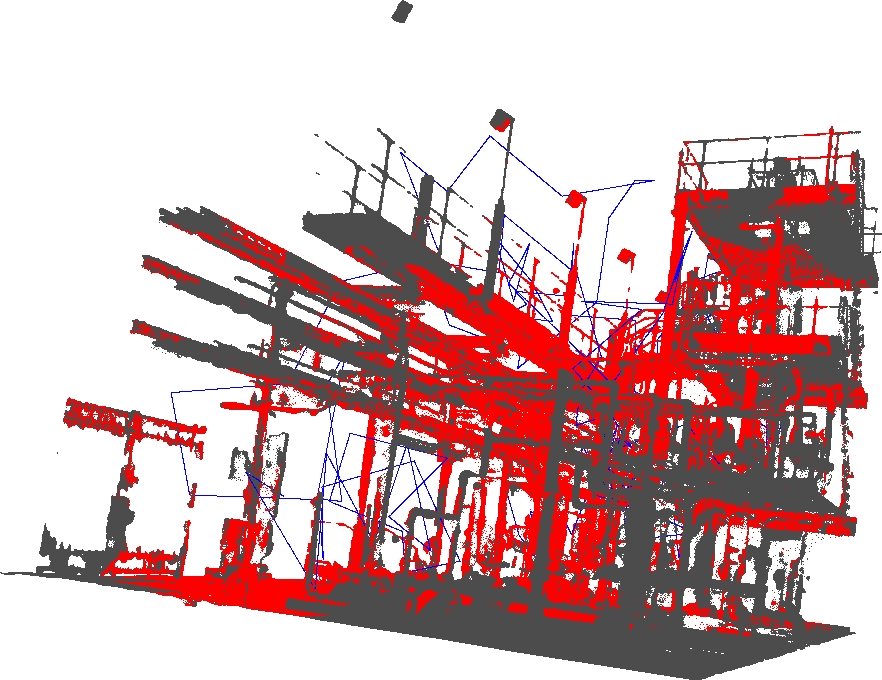} \includegraphics[width=0.23\textwidth]{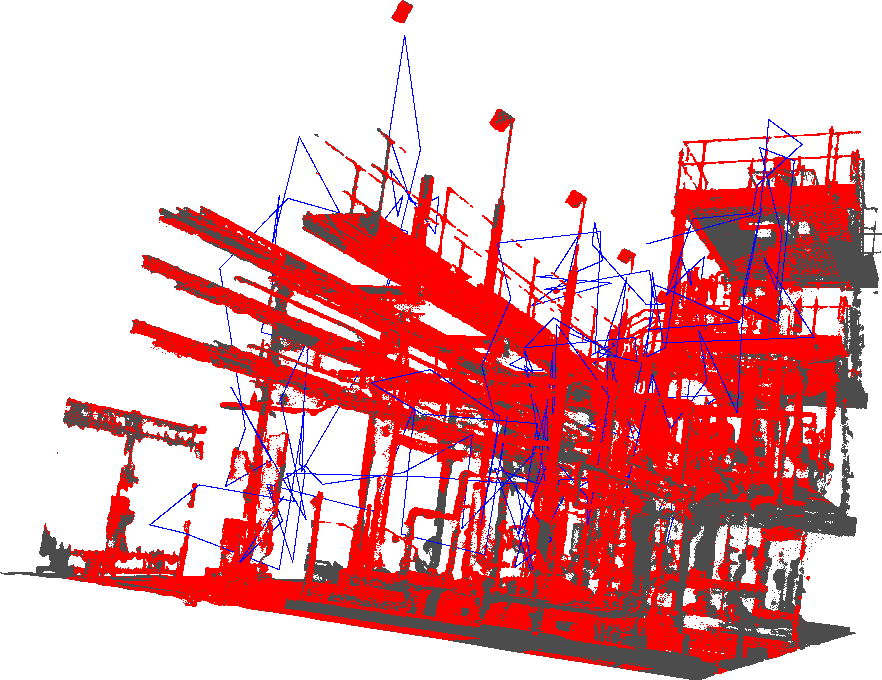}} 
     


     }
  \caption{Sample trajectories in \textbf{Plant-s} planned by \tpipim, with red covered, black uncovered and blue the inspection path. \tpipim plans within 2 hours (Up) and 3 hours (Bottom), with  $55\%$ and $62\%$ coverage. \label{sample-traj}}
  \vspace{-1.5em}
\end{figure}

 \textbf{2) Comparison between \tpipim, \iris and \irism:} 
 We ran \iris in both scenarios for 24 hours without getting any results. Therefore, for the rest of the comparison, we use \irism. 
 In all scenarios, \tpipim achieves similar coverage results as \irism. \irism has asymptotically optimal guarantees, achieved by searching on an RRG. Despite its only 2 GB baseline memory cost, \irism searches numerous number of inspection paths, resulting in huge total memory cost. As shown in Table. \ref{tbl: all}, in \textbf{Plant-s} and \textbf{Plant}, out of 10 runs, \irism ran out of memory for 6 and 10 runs with 96 GB total memory limit. In instances where \irism exhausts its memory, we capture and save the best inspection path  \irism generated up to the point before the process terminated. On the other hand, \tpipim did not fail any of the runs, with only a 1 GB total memory limit. One might view the comparison unfair because \tpipim does not have any optimality guarantees. However, by just comparing the baseline memory consumptions, which always exists regardless of whether a search for optimality is performed, \ipim also brings in $\sim 25\times$ and $\sim 70\times$ memory efficiency in \textbf{Plant-s} and \textbf{Plant}.


\fref{sample-traj} provides a visualization of the coverage of \tpipim for 2 hours and 3 hours planning time. The inspection path demonstrates that, as time increases, \ipim enables the planner to navigate through a cluttered area (see bottom right) to achieve better coverage.

\section{Summary \label{sec:con}}

We present a set of primitive computations, called \ipim, to allow sampling-based inspection planning to efficiently use implicit environment models. Evaluation indicates that \ipim substantially reduces the memory cost in inspection planning of large, cluttered, and confined environments. Many avenues are possible for future work. For instance, can better efficiency be gained with other implicit environment models? And, how to incorporate uncertainty in the environment and perception?

\section*{ACKNOWLEDGMENT}
We thank Abyss Solutions for collecting and providing the San Jacinto dataset. This work was supported by the ARC Research Hub in Intelligent Robotic Systems for Real-Time Asset Management (IH210100030).

\bibliographystyle{IEEEtran}
\bibliography{IEEEabrv, reference}

\end{document}